%% file: main.tex
\documentclass[conference]{IEEEtran}

\usepackage{amsmath}
\usepackage{cite}
\usepackage{graphicx}
\usepackage{epstopdf} %converting to PDF
\usepackage{multirow}
\usepackage{pbox}
\usepackage{color}
\usepackage{xcolor}
\usepackage[official]{eurosym}
\usepackage[nolist]{acronym} 
\usepackage{amsmath}
\usepackage{algorithm}
\usepackage[noend]{algpseudocode}
\usepackage{comment}

\title{Reinforcement Learning on Computational Resource Allocation of Cloud-based Wireless Networks} 
\author{  
Beiran Chen\IEEEauthorrefmark{1}, Yi Zhang\IEEEauthorrefmark{1}, George Iosifidis\IEEEauthorrefmark{1}, and 
Mingming Liu\IEEEauthorrefmark{2}\\ 
\IEEEauthorblockA{\IEEEauthorrefmark{1} CONNECT centre, Trinity College Dublin, Ireland, \{chenbe, zhangy8, george.iosifidis\}@tcd.ie}
\IEEEauthorblockA{\IEEEauthorrefmark{2} Insight Centre for Data Analytics, Dublin City University, Ireland, mingming.liu@dcu.ie}
}

\begin{document}

\maketitle
\begin{abstract}

Wireless networks used for Internet of Things (IoT) are expected to largely involve cloud-based computing and processing. Softwarised and centralised signal processing and network switching in the cloud enables flexible network control and management.
In a cloud environment, dynamic computational resource allocation is essential to save energy while maintaining the performance of the processes. The stochastic features of the Central Processing Unit (CPU) load variation as well as the possible complex parallelisation situations of the cloud processes makes the dynamic resource allocation an interesting research challenge. This paper models this dynamic computational resource allocation problem into a Markov Decision Process (MDP) and designs a model-based reinforcement-learning agent to optimise the dynamic resource allocation of the CPU usage.
Value iteration method is used for the reinforcement-learning agent to pick up the optimal policy during the MDP. To evaluate our performance we analyse two types of processes that can be used in the cloud-based IoT networks with different levels of parallelisation capabilities, i.e., Software-Defined Radio (SDR) and Software-Defined Networking (SDN).
The results show that our agent rapidly converges to the optimal policy, stably performs in different parameter settings, outperforms or at least equally performs compared to a baseline algorithm in energy savings for different scenarios. 

\end{abstract}

\begin{IEEEkeywords}
Reinforcement learning, IoT, Cloud, SDN, SDR, Markov Decision Process
\end{IEEEkeywords}

% ================================================================
\input{introduction}
\input{background}

\input{system_model}

\input{results}

\input{conclusion}

\input{acro_list}
% ================================================================

\section*{Acknowledgement}
This publication has emanated from the research supported
by the European Commission Horizon 2020 under grant agreements
no. 732174 (ORCA), and co-funded
under the European Regional Development Fund from
Science Foundation Ireland under Grant Number 13/RC/2077
(CONNECT). This material is also based on works supported by Science Foundation Ireland under Grant No. SFI/12/RC/2289\_P2.

\bibliographystyle{IEEEtran}
\bibliography{bibliography}
\end{document}

%% file: introduction.tex
\section{Introduction}
\label{sec:introduction}

%Wireless networks for \ac{IoT} are often associated with low latency, high reliability, flexible control and high energy efficiency.
%Cloud computing is promising to be integrated with next-generation \ac{IoT} networks to achieve these requirements \cite{BiGi14}. 
Cloud-based systems have been proposed and implemented for many promising \ac{IoT} applications, such as smart city and e-health~\cite{NaGu18, MoRoSi18}. Two enabling technologies, \ac{SDR} and \ac{SDN}, recently started to be utilised in the cloud to support the virtualisation of the \ac{IoT} network~\cite{AhAlMa19}.
In a cloud environment, computational resources (e.g. CPU, memory, and storage) are virtualised and allocated to fulfil the requirements
of different \ac{IoT} services.
Optimising computational resource allocation in the cloud to balance energy consumption and processing performance is an important research topic. 
%One recently developed and enhanced artificial intelligent technology, namely 
Reinforcement learning provides an effective method for solving such an optimisation problem in a stochastic and dynamic environment. 

In this paper, we design and implement a model-based reinforcement-learning agent based on \ac{MDP}, to intelligently allocate  computational resources for cloud-based wireless networks. The agent analyses the state transition probabilities between different states, reflecting different levels of \ac{CPU} utilisation,  as well as the corresponding reward functions, and then decides to either remain in the same state or to change states by adding/reducing one CPU core. The reward function for each state-action pair drives the agent to optimise the total reward after multiple \ac{MDP} steps.
We use a value iteration algorithm  \cite{suttonbook} for the agent to solve this \ac{MDP} and arrive at an optimal policy 
%out of the state-action pairs which
that balances the energy savings and cloud performance.

We simulate many different \ac{SDR} and \ac{SDN} scenarios 
%for both \ac{SDR} and \ac{SDN} cases 
to evaluate the performance of our reinforcement-learning agent. We have found that the optimal policies carried out by the agent are different for \ac{SDR} or \ac{SDN} cases. The results show that our agent with value iteration outperforms a baseline algorithm in terms of energy savings in some of the scenarios while achieving similar performance in the other scenarios. We also investigate the performance of our agent with different transition probabilities as well as a higher number of \ac{MDP} states to verify the scalability of our agent. 

Our contribution in this paper can be summarised as follows: 1) the modelling of the dynamic CPU resource allocation problem in the cloud-based wireless networks by \ac{MDP}; 2) the utilisation of value iteration method to solve the \ac{MDP} to get the optimal policy; and 3) the comprehensive analysis of the performance of our reinforcement learning agent in various scenarios with different parameters. 

%% file: background.tex
\section{Background and related work}
\label{sec:background}

Reinforcement learning was originally invented for robotics and automation.
Recently with the artificial intelligence and machine learning being introduced to broader research areas, reinforcement learning has started to be applied in communications and networking. The main advantage of reinforcement learning is that it does not need a large dataset during the training process. A reinforcement-learning agent is capable to sense the environment and learn to make decisions by itself during the training. To design a reinforcement learning agent, a description of the environment is essential, either mathematically modelled as an input to the agent (i.e., model-based), or learnt by the agent itself (i.e., model-free). In this paper we model the cloud computational resource allocation problem as an \ac{MDP}, which is a commonly used model to describe a stochastic process \cite{suttonbook}. We then get optimal solutions for the agent by value iteration, which is an effective reinforcement-learning algorithm to solve the \ac{MDP}.

There are some recent works published on reinforcement learning utilisation on the computational resource allocation and offloading at the edge computing or fog computing. For instance, authors in \cite{LiGaLv18} used reinforcement learning to optimise the computational task offloading problem from mobile users to edge computing servers. The authors in \cite{LeTh18} proposed a reinforcement-learning method to offload the computational task from one user to another user in ad-hoc wireless networks with \ac{MDP} model.
Besides, authors in \cite{TaZhZh18} applied \ac{MDP} to model the container migration problem in fog computing and then use reinforcement learning to design an agent that migrates containers between different physical servers while optimising the total power consumption.
Despite the fact that reinforcement learning has been used for addressing many different resource or task allocation problems in cloud computing, to the best of our knowledge, our work is the first one that investigates the problem of dynamic CPU resource allocation of cloud-based wireless networks by reinforcement learning.

%% file: system_model.tex
\section{System Model}
\label{sec:system_model}

In this section we first describe the \ac{MDP} model that we build for the dynamic computational resource allocation problem of cloud-based wireless networks. After that we define the reward function, value function and Q function for this \ac{MDP} model, as the parameters for our reinforcement-learning agent. Finally we present the value iteration algorithm that can help solve the \ac{MDP} model and get optimal policies by reinforcement learning.

\subsection{The MDP model}

%The utilisation percentage for CPU cores in cloud-based networking systems varies over time depending on the processes running on the system. 
%Some of the processes can be parallelised into multiple cores while some of them cannot, therefore the utilisation percentage differs over different cores in the multi-core cloud system. 
%Our reinforcement-learning agent is designed to dynamically add or reduce number of CPU cores assigned to these processes according to the demands, in order to save energy while avoiding overloading of the CPU cores. Our reinforcement-learning agent is model-based, meaning that the variation of percentages of CPU loads and number of assigned CPUs can be modelled as an \ac{MDP} process.

\ac{MDP} is a discrete-time stochastic process that is used for modelling the decision making procedure involving multiple states and actions in a stochastic environment \cite{Bel}. 
According to our observation of the statistics of Trinity College Dublin's cloud-based Iris testbed \cite{iris_link}, we model the stochastic process of the CPU usage and the CPU allocation of cloud processes in our testbed as an \ac{MDP}. Supposing $L_{max}$ denotes the maximum instant CPU load percentage for the cores that are being used (e.g., if a container in the cloud is using 3 CPU cores with the load percentage of 50\%, 60\% and 70\%, $L_{max}=70\%$), the core with $L_{max}$ becomes the bottle neck of the container in terms of the processing performance.
We then categorise the CPU load percentage $L_{max}$ into three levels, i.e., low utilisation state $s0$ ($<20\%$), medium utilisation state $s1$ ($20-80\%$) and high utilisation state $s2$ ($>80\%$), corresponding to the three states in the \ac{MDP}.
Note that we have also made the cases with more states (i.e., more fine-grained categorisation of CPU load percentage levels) later on in the result section and proved that the results with more states are similar with the three states case.

After defining the states, we define the three options of actions for the reinforcement-learning agent in the \ac{MDP} to be 1) keeping the current CPU numbers, i.e., $a0$; 2) adding one CPU core, i.e., $a1$; 3) reducing one CPU core, i.e., $a2$. The transitions between the states $s$'s after operating the action $a$'s are stochastic processes due to the uncertainty of the parallelisation situations of the processes running on the CPU cores, i.e., adding or reducing one CPU core would not necessarily change the states.
%For example, on one hand, adding one CPU core would not necessarily decrease the $L_max$ if the process causing the $L_max$ can not be parallelised. On the other hand, reducing one CPU core would not necessarily increase the $L_max$ in the case that the processes that were running on the deducted CPU core are moved and added to the core that has the new load still below $L_max$ after the CPU core reduction. Therefore adding or reducing CPU cores would not necessarily change the states. 
We define the state transition probability $P(s'|s, a)$ as the probability of getting into the next state $s'$ (either the same one or a different one) if taking action $a$ from state $s$. We also define the reward of this transition as $r(s, a, s')$, as the parameter for our reinforcement learning agent. 
The time interval of the agent taking the actions can be customised to be different levels (e.g. in seconds, in minutes or in hours) depending on the needs of the network administrators.

\begin{figure}[ht]
    \vspace{-0.1in}
	\centering
\includegraphics[width=0.5\textwidth]{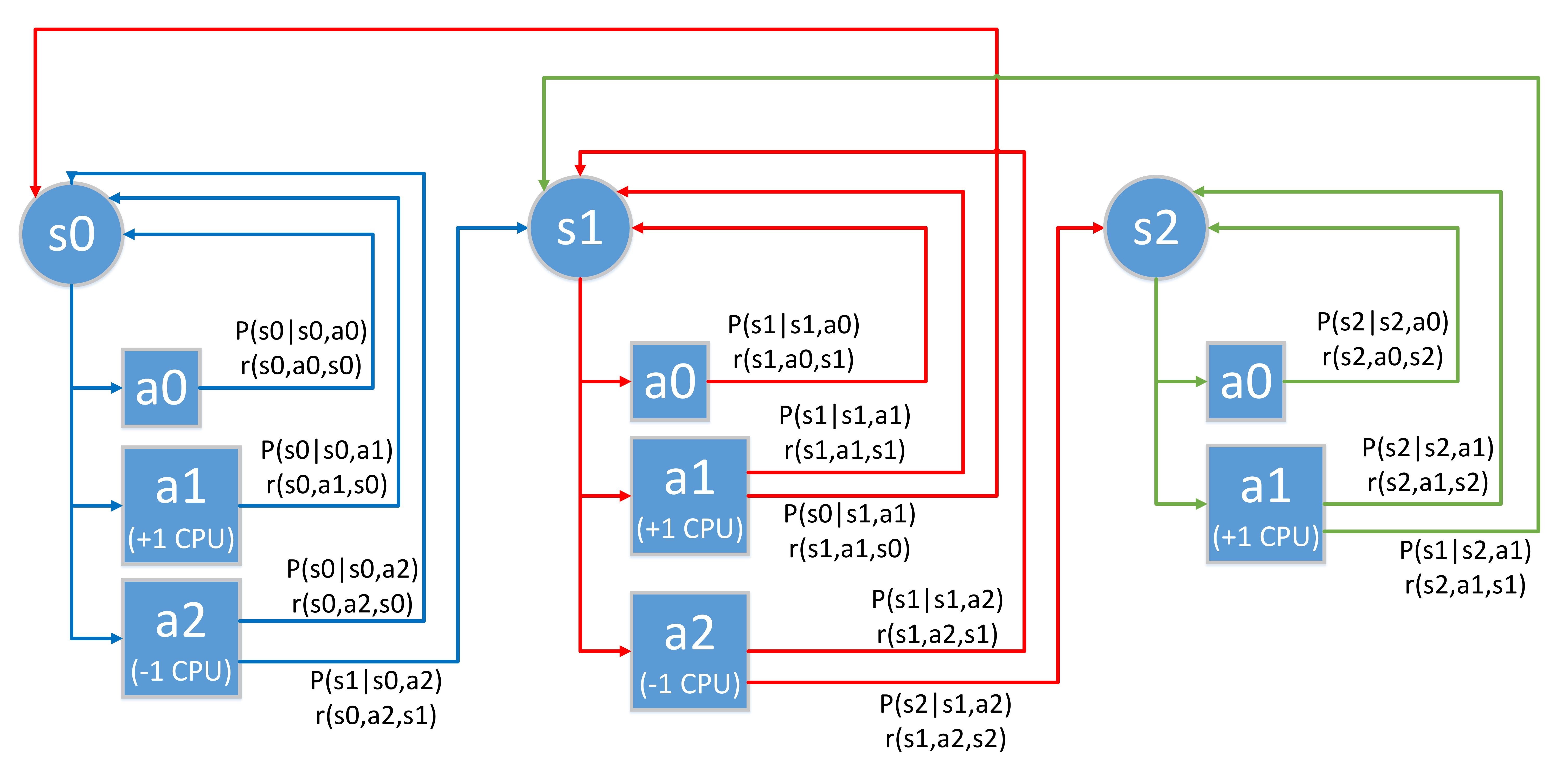}
	\caption{MDP state transition diagram}
	\label{fig:MDP_picture}
    \vspace{-0.1in}
\end{figure}

Figure \ref{fig:MDP_picture} illustrates the state transition diagram of the aforementioned \ac{MDP} model, including the 3 states, i.e. $s0$, (low CPU load percentage), $s1$ (medium CPU load percentage) and $s2$ (high CPU load percentage), as well as the 3 actions, i.e., $a0$ (doing nothing), $a1$ (adding one CPU core) and $a2$ (reducing one CPU core). The figure also shows the transition probabilities $P(s'|s, a)$ as well as the corresponding rewards $r(s, a, s')$ with the arrowed lines between the states and actions. 
Different colours of arrowed lines are only for improving visibility.
Note that taking action $a2$ from state $s2$ is not considered because it is not realistic to reduce one CPU core when the CPU load percentage is already high.

\subsection{Reward function, Value function, and Q function}

In this subsection we define the reward function, value function and Q function from this \ac{MDP} for the reinforcement learning~\cite{suttonbook}.
The objective of solving this \ac{MDP} is to find a policy $\pi$ that maximises the total reward $R$ when the system transits between the states after certain \ac{MDP} steps $T$. The total reward $R$ follows Equation (\ref{eqn:total_rewards}) where $r_i$ denotes the reward of each time step $i$ and $\gamma$ is the discount factor ($0 < \gamma < 1$, set to be 0.9 in this paper) that avoids the total reward going to infinity~\cite{suttonbook}.

\begin{equation}
    R = \sum_{i=1}^{T}\gamma^{i-1}r_i
    \label{eqn:total_rewards}
\end{equation}

We then define the value function of each state $V^{\pi}(s)$, denoting the expected total reward for an agent starting from state $s$ with the policy $\pi$ (shown in Equation (\ref{eqn:value_function})).  In this way, $V^{\pi}(s)$ indicates how good the state $s$ is for an agent to stay. Among all policy $\pi$'s, there existing an optimal policy $\pi^*$ that makes $V^{\pi}(s)$ to be maximum (shown in Equation (\ref{eqn:optimal_pi}))~\cite{suttonbook}.

\begin{equation}
    V^{\pi}(s) = E[\sum_{i=1}^{T}\gamma^{i-1}r_i]
    \label{eqn:value_function}
\end{equation}

\begin{equation}
    \pi^* = \arg \max_{\pi} V^{\pi}(s)
    \label{eqn:optimal_pi}
\end{equation}

We then define $V(s)$ as the short version for $V^{\pi}(s)$ from now on for simplicity.
In order to get the optimal $V(s)$, in reinforcement learning, the agent needs to try all policy $\pi$'s that include all possible combinations of state-action pairs $(s, a)$. The Q function $Q(s, a)$ for each state-action pair can be defined to indicate how beneficial it is for the agent to use action $a$ when in the state $s$. Therefore the maximum value of $V(s)$ ($V^*(s)$) equals to the maximum value of $Q(s, a)$ ($Q^*(s, a)$) for all the possible action $a$'s (shown in Equation (\ref{eqn:optimal_q}))~\cite{suttonbook}.

\begin{equation}
    V^*(s) = Q^*(s, a) = \max_{a}Q(s, a)
    \label{eqn:optimal_q}
\end{equation}

According to the Bellman Equation \cite{Bellmanbook2003}, The optimal $Q^*(s, a)$ equals to the summation of 1) the expectation of immediate reward $r(s, a, s')$ after taking action $a$ from state $s$ (considering all possible next states $s'$'s) and 2) the discounted expectation of all future maximum rewards $V^*(s')$ (for all possible next states $s'$'s as well), shown in Equation (\ref{eqn:bellman})~\cite{suttonbook}.

\begin{equation}
    Q^*(s, a) = \sum_{s'}P(s'|s, a)r(s, a, s') + \gamma \sum_{s'}P(s'|s, a)V^*(s')
    \label{eqn:bellman}
\end{equation}

Therefore, according to Equation (\ref{eqn:optimal_q}) and (\ref{eqn:bellman}), we can derive Equation (\ref{eqn:optimal_v_and_q}), for all possible state $s'$'s transiting from state $s$ taking action $a$~\cite{suttonbook}.

\begin{equation}
    V^*(s) = \max_{a}\sum_{s'}P(s'|s, a)(r(s, a, s') + \gamma V^*(s'))
    \label{eqn:optimal_v_and_q}
\end{equation}

\subsection{Reinforcement learning and value iteration}

According to the aforementioned Equations (\ref{eqn:value_function})-(\ref{eqn:optimal_v_and_q}), the reinforcement learning agent needs to obtain the maximum $V^*(s)$ and the corresponding optimal policy $\pi^*$ to perform an optimal solution to this \ac{MDP}. Assuming $P(s'|s, a)$ and $r(s, a, s')$ are both known for all the triple tuples $(s, a, s')$, we can use a reinforcement learning method, namely value iteration algorithm, to solve the optimal $V^*(s)$, by a recursive method using Equation (\ref{eqn:optimal_v_and_q}). The pseudo code of the value iteration algorithm can be expressed as the following Algorithm \ref{alg:vi} \cite{suttonbook}:

\begin{algorithm}
\caption{Value iteration algorithm}\label{alg:vi}
\begin{algorithmic}[1]
\State Initialisation: initial $V_{(0)}(s) = 0$ for all $s$
\For {$j = 0, 1, 2 ...$ ($j$ denotes the iteration step)}
\For {all $s$ in state set $S$}
\State $Q_{j}(s, a) \gets \sum_{s'}P(s'|s, a)(r(s, a, s') + \gamma V_{j}(s'))$
\State $V_{j+1}(s) \gets \max_{a} Q_{j}(s, a)$
\If {$|V_{j}(s) - V_{j+1}(s)| < \epsilon $}
\State \textbf{break}
\EndIf
\EndFor
\EndFor
\State $V^*(s) \gets V_{j}(s)$
\State \textbf{return} $V^*(s)$ and $\pi^*$ 
\end{algorithmic}
\end{algorithm}

The value iteration algorithm first initialises the value function $V(s)$ with arbitrary values (in our case, zeros), and then updates it with the value of the latest Q function $Q(s, a)$ when making a step ahead. After a number of iterations the value function $V(s)$ converges to the optimal value $V^*(s)$, when the difference between the last two iterations is less than a very small value $\epsilon$. The corresponding policy $\pi^*$ becomes the optimal policy (i.e., the optimal state-action pairs $(s, a)$).

%% file: results.tex
\section{Experimental results}
\label{sec:results}

To evaluate the performance of our agent running the aforementioned value iteration algorithm in a cloud-based wireless network scenario, we investigate two commonly-used network processes, i.e., \ac{SDN} and \ac{SDR}. We conduct our simulations with different experimentation parameters and compare the results with another baseline algorithm. 

\subsection{Definition of transition probabilities and rewards of \ac{MDP}}

We start with an experiment of 3 states for the \ac{MDP} (shown in Figure \ref{fig:MDP_picture}).
We define the parameters we use for the \ac{MDP} simulation for both \ac{SDN} and \ac{SDR} cases in Table \ref{tab:parameters_3_states}. The transition probabilities between the 3 states (i.e., 3 levels of CPU core usage percentage) are derived from
the general experimental data statistics of the Trinity College Dublin Iris testbed \cite{iris_link}. 
The \ac{SDN} and \ac{SDR} cases have different transition probabilities, due to the facts that 1) a single \ac{SDN} process utilises a lower percentage of a CPU core but there are usually multiple processes running at the same time to share the same CPU core, meaning that \ac{SDN} processes are more likely to be parallelised; 2) a single \ac{SDR} process utilises a higher percentage of a CPU core and is usually not capable to be parallelised. Therefore, the probability for \ac{SDN} to transit to other states is larger than the one for \ac{SDR} when increasing or decreasing the number of allocated CPU cores.

Besides, we define the normalised rewards of the reinforcement learning procedure when solving the \ac{MDP} (also shown in Table \ref{tab:parameters_3_states}). The definition of the rewards is based on the objective that is to minimise the total number of running CPU cores (to save energy) while avoiding long-term high CPU load percentage (to guarantee the overall performance of the cloud system). 
In general, the reward is positive when 1) reducing one CPU core and the CPU load keeps the same level; or 2) adding one CPU core and the CPU load level gets lower. The reward is negative when 1) doing nothing (i.e. with action $a0$); or 2) adding one CPU core but CPU load level remains; or 3) reducing one CPU core and CPU load level increases.

\begin{table}

\centering
\caption{State transition probability and corresponding rewards for the 3 states transition map}
\label{tab:parameters_3_states}
\begin{tabular}{|c|c|c||c|c|}
\hline\hline
 \multicolumn{3}{|c||}{Transition probabilities} & \multicolumn{2}{|c|}{Rewards} \\ \hline
 & SDN & SDR &   &  SDN \& SDR  \\\hline\hline
$P(s0 |s0, a0)$ & 1 & 1 & $r(s0, a0, s0)$ & -3 \\ \hline
$P(s0 |s0, a1)$ & 1 & 1 & $r(s0, a1, s0)$ & -5\\ \hline
$P(s0 |s0, a2)$ & 0.3 & 0.7 & $r(s0, a2, s0)$ & +5\\ \hline
$P(s1 |s0, a2)$ & 0.7 & 0.3 & $r(s0, a2, s1)$ & -5\\ \hline
$P(s1 |s1, a0)$ & 1 & 1 & $r(s1, a0, s1)$ & -1\\ \hline
$P(s0 |s1, a1)$ & 0.8 & 0.2 & $r(s1, a1, s0)$ & +5\\ \hline
$P(s1 |s1, a1)$ & 0.2 & 0.8 & $r(s1, a1, s1)$ & -5\\ \hline
$P(s1 |s1, a2)$ & 0.3 & 0.7 & $r(s1, a2, s1)$ & +5\\ \hline
$P(s2 |s1, a2)$ & 0.7 & 0.3 & $r(s1, a2, s2)$ & -5\\ \hline
$P(s2 |s2, a0)$ & 1 & 1 & $r(s2, a0, s2)$ & -5\\ \hline
$P(s1 |s2, a1)$ & 0.8 & 0.2 & $r(s2, a1, s1)$ & +5\\ \hline
$P(s2 |s2, a1)$ & 0.2 & 0.8 & $r(s2, a1, s2)$ & -5\\ \hline
\hline
\end{tabular}
\vspace{-0.1in}
\end{table}

\subsection{Results with 3 states, certain transition probability and predefined rewards}

As mentioned in the previous section, we design the reinforcement-learning agent to find the optimal policy $\pi^*$ using the value iteration algorithm (Algorithm \ref{alg:vi}). We use Python 3.5 open-source \ac{MDP} library \cite{mdp_link} to implement the \ac{MDP} model, the value iteration procedure and simulation of the results.
The threshold of the convergence ($\epsilon$ in the Algorithm~\ref{alg:vi}) is set to be a very small value ($10^{-3}$). The value iteration algorithm converges rapidly to optimal values. For SDN case it converges after 66 iterations, while for SDR case it converges after 56 iterations.

\begin{figure}[ht]
    \vspace{-0.1in}
	\centering
\includegraphics[width=0.45\textwidth]{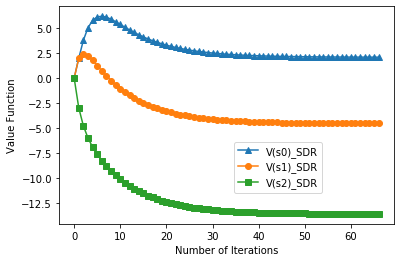}
	\caption{Value function of the 3 different states vs. Iterations for SDR}
	\label{fig:value_function_SDR}
    \vspace{-0.1in}
\end{figure}

\begin{figure}[ht]

	\centering
\includegraphics[width=0.45\textwidth]{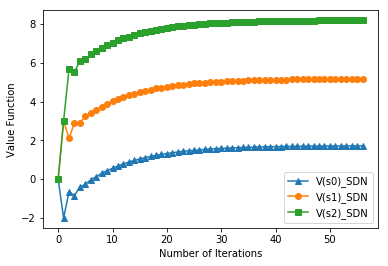}
	\caption{Value function of the 3 different states vs. Iterations for SDN}
	\label{fig:value_function_SDN}
    \vspace{-0.1in}
\end{figure}

Figure \ref{fig:value_function_SDR} and Figure \ref{fig:value_function_SDN} show the value function variations and convergence during the value iteration process for \ac{SDR} and \ac{SDN} respectively. For the case of \ac{SDR} there is $V(s0)>V(s1)>V(s2)$ when converged; while for the case of \ac{SDN} there is $V(s2)>V(s1)>V(s0)$ when converged. The results mean that in the case of \ac{SDR} the agent prefers to go for a lower CPU load state, while in the case of \ac{SDN} the agent prefers to go for a higher CPU load state. To understand the decision of the agent, we should take into account the fact that the \ac{SDN} processes are easier to be parallelised than the \ac{SDR} processes, therefore it is safer to push to a higher CPU load state for \ac{SDN} to save resources than \ac{SDR}, because in the case of \ac{SDN} adding one CPU will more likely bring down the high CPU load and get positive reward. However for the \ac{SDR} case the states tend to stay when adding/reducing CPUs therefore there is a higher risk for the agent to stay in the high CPU load state and get continuously high penalties.

Besides, the agent also derives the optimal policy denoted by state-action pairs $(s,a)$. For the case of \ac{SDR} they are $(s0,a2)$, $(s1,a2)$, and $(s2,a1)$, while for the case of \ac{SDN} they are $(s0,a2)$, $(s1,a1)$, and $(s2,a1)$. These two cases have the same optimal actions at the state $s0$ and $s2$, which are $a2$ and $a1$ respectively, meaning that the agent chooses to reduce one CPU when the CPU load is low and to add one CPU when the CPU load is high, which is straightforward to understand since these two actions are the only options in these two states to get an immediate positive reward.
However, the optimal action for the medium state $s1$ is different for \ac{SDR} and \ac{SDN}, For the case of \ac{SDR} the agent prefers to reduce one CPU at the state of $s1$ while for the case of \ac{SDN} the agent prefers to add one CPU at the state of $s1$. These two decisions are not obvious to understand since both actions on state $s1$ can make the same immediate reward. Therefore the agent has to decide these actions not only by the immediate reward, but also by maximising the total reward considering future rewards with the discount factor $\gamma$ (see Equation (\ref{eqn:total_rewards})). These decisions are optimised by the value iteration.

\subsection{Results with different transition probabilities}

In the previous subsection, we obtain the transition probabilities shown in Table \ref{tab:parameters_3_states} based on the general statistics of the Trinity College Dublin Iris testbed \cite{iris_link}. 
%These parameter settings align with the different features of process parallelisation in \ac{SDR} and \ac{SDN}, as mentioned in previous sections. 
In this subsection we change these probabilities and see how the changes would effect to the results, to further evaluate the scalability of our algorithm to other testbeds/scenarios. We keep those probability values that equal to 1 in Table \ref{tab:parameters_3_states} for the cases in which state transition is definite, and vary the other "not-equal-to-1" values that are for the transition probability between different states. We assume that
for the \ac{SDR} case those values vary among 0.01, 0.1, 0.2, 0.3, and 0.4, while for the \ac{SDN} case those values vary among 0.6, 0.7, 0.8, 0.9 and 0.99. The reason is that \ac{SDN} processes are more likely to be parallelised thus more likely to transit to another state when adding/reducing one CPU core. 

\begin{figure}[ht]
    \vspace{-0.1in}
	\centering
\includegraphics[width=0.45\textwidth]{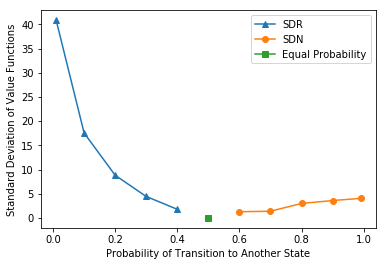}
	\caption{Standard deviation of the value functions in different transition probability cases}
	\label{fig:different_probability_standard_deviation}
	\vspace{-0.1in}
\end{figure}

The results in Figure \ref{fig:different_probability_standard_deviation} show the relation between the aforementioned state transition probability from one state to another state (X axis) and the standard deviation of the value functions of the 3 states (Y axis). 
The results show that the variation of transition probabilities affects a lot to the standard deviation of the value functions for the \ac{SDR} case but not so much for the \ac{SDN} case. Besides, the curves for the \ac{SDR} and \ac{SDN} cases show different tendencies with the increasing of transition probability. In general, higher standard deviation of value functions means the reinforcement-learning agent has more obvious preferences on different states.
%For the \ac{SDR} case the standard deviation of the 3 state value functions varies from 1 to 40 with the changes of transition probabilities, while for the \ac{SDN} case the standard deviation only varies between 0 and 5. 
In Figure \ref{fig:different_probability_standard_deviation} we also plot the case for equal probability values (i.e. with 0.5 transition probability, meaning equal probability to stay in the same state or to transit to another state), in which case the state value functions all equal to 0 ($V(s0)=V(s1)=V(s2)=0$). This means that the agent is indifferent to the 3 different states therefore the value iteration does not work properly in this case.
%as mentioned earlier in this subsection.

%The transition probability affects more in the \ac{SDR} case than in the \ac{SDN} case is not obvious to understand. However, the fact that the standard deviation of the state value functions is generally lower in higher transition probability cases (i.e., \ac{SDN} cases) than in lower transition probability cases (i.e. \ac{SDR} cases) is understandable. It is because that when transition probability is high, it is relatively easy for the agent to change states by just adding or reducing one CPU, so the benefit of starting from which state does not matter that much.

\subsection{Results compared to other algorithms}

In this subsection, we compare the results of our value iteration algorithm with another baseline algorithm. 
According to the previous analysis and review values in Table \ref{tab:parameters_3_states}, for the states $s0$ or $s2$ there is only one optimal choice of state-action pair, (i.e., $(s0, a2)$ or $(s2, a1)$), which has positive reward value. Therefore any optimisation algorithm would choose these two state-action pairs. 
However, the difference occurs when transiting from the state $s1$, 
where there is no obvious preference for the agent on taking the action $a1$ or $a2$ in terms of immediate rewards. The benefit of the algorithm will appear with time goes by. Therefore, to compare the performance of our algorithm and the baseline algorithm, we simulate the \ac{MDP} procedure to calculate the cumulative reward (defined by Equation \ref{eqn:total_rewards}) after a certain number of \ac{MDP} steps.

\begin{figure}[ht]
  
	\centering
\includegraphics[width=0.45\textwidth]{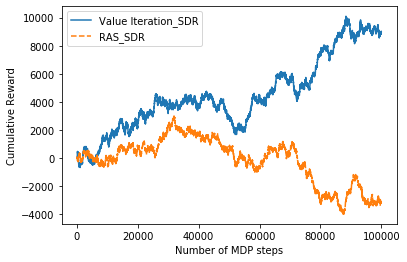}
	\caption{Cumulative reward comparison between value iteration approach and RAS approach for an SDR MDP case}
	\label{fig:rewards_comparison_SDR}
    \vspace{-0.1in}
\end{figure}

We define the baseline algorithm, namely \ac{RAS} algorithm, to compare with our value iteration algorithm. The \ac{RAS} algorithm randomly chooses the action $a1$ or $a2$ made at the state $s1$ with equally 50\% probability respectively. Figure \ref{fig:rewards_comparison_SDR} shows the comparison of cumulative reward between our value iteration approach and the \ac{RAS} approach for one single time-based \ac{MDP} simulation. The simulation is conducted under the same SDR case with the state transition probability equal to 0.1 for both adding one CPU and reducing one CPU. The results show that our agent with value iteration gets more cumulative reward during the reinforcement learning process than the agent with \ac{RAS}, meaning our agent works more efficiently to reach our goal.

We also build simulations to compare the energy savings and CPU working performance for value iteration and \ac{RAS} algorithm, in a cloud with 1000+ available CPU cores.
Instead of showing the results for a single \ac{MDP} as Figure \ref{fig:rewards_comparison_SDR}, we build 10000 independent \ac{MDP} simulations with each simulation having 1000 \ac{MDP} steps and take average results to get more reliable comparisons.

\begin{figure}[ht]

	\centering
\includegraphics[width=0.45\textwidth]{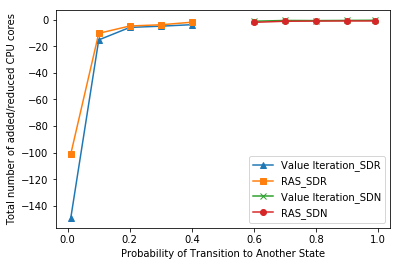}
	\caption{Comparison between value iteration and RAS for total number of added/reduced CPUs in different transition probability cases}
	\label{fig:CPU_comparison}
    \vspace{-0.1in}
\end{figure}

\begin{figure}[ht]

    \centering
\includegraphics[width=0.45\textwidth]{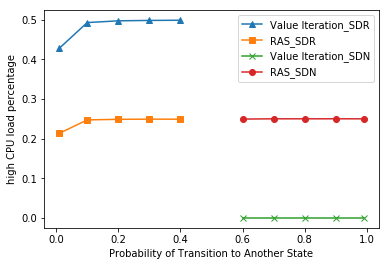}
	\caption{Comparison between value iteration and RAS for percentage of time in high CPU load state ($s2$) in different transition probability cases}
	\label{fig:high_percentage_comparison}
    \vspace{-0.1in}
\end{figure}

To showcase the energy savings, Figure \ref{fig:CPU_comparison} shows the comparison between value iteration and \ac{RAS} for total number of added/reduced CPU cores in different transition probability cases (positive numbers on Y-axis mean adding CPUs while negative numbers mean reducing CPUs). Besides, to showcase the performance maintenance, Figure \ref{fig:high_percentage_comparison} shows the comparison between value iteration and \ac{RAS} for percentage of time in high CPU load state ($s2$) in different transition probability cases. 

%These simulations are also based on the average of 10000 independent \ac{MDP} experiments, each of which has 1000 \ac{MDP} steps.

By looking at Figure \ref{fig:CPU_comparison},
for \ac{SDR} case, our agent with value iteration reduces more number of CPU cores in the MDP than \ac{RAS}, especially for cases where transition probability is less than 0.2. This means that our agent saves more energy than the baseline \ac{RAS} in the cloud systems for the \ac{SDR} case. For instance, for the case where transition probability equals to 0.01, our value iteration agent saves energy consumption of 40 more CPUs than \ac{RAS} does. If one high-performance CPU core consumes 80 watts \cite{intel_link}, saving 40 CPUs corresponds to around 76.8 kWh energy savings per day.
However, Figure \ref{fig:high_percentage_comparison} shows that
value iteration has around 25\% more time in high-load CPU state than \ac{RAS} in \ac{SDR} cases (although it is still below 50\% of time), meaning that our agent takes higher risk on keeping the CPUs on high-load states than \ac{RAS}.
For \ac{SDN} case, our agent with value iteration has similar CPU core reductions (i.e. energy savings) as \ac{RAS}, but takes lower risk on keeping the CPUs on high-load states than \ac{RAS}. 
Note that we don't consider the case in which the transition probability equals to 0.5 on the X-axis because in this case the value iteration does not work properly (as mentioned in Figure~\ref{fig:different_probability_standard_deviation}).

\subsection{Results with more states}

%The previous subsections have shown the results for 3-state transition \ac{MDP}, corresponding to the \ac{MDP} state transition diagram in Figure \ref{fig:MDP_picture}. 
In this subsection, we investigate the case with more \ac{MDP} states (4 states) and see if the results follow the similar trend, to evaluate the scalability of our agent using value iteration when the CPU-usage levels are defined to be more fine-grained.
%Figure \ref{fig:value_function_SDR_4_states} and Figure \ref{fig:value_function_SDN_4_states} show the value function results for the case of 4-state transition for \ac{SDR} and \ac{SDN} respectively. 
To make the case for ``4 states"
%we mean that we divide utilisation level of the CPU percentage into 4 levels (for example, 0-10\%, 10-50\%, 50-90\%, and 90-100\%, denoted as $s0$, $s1$, $s2$, and $s3$), 
we add one more intermediate state to the previous 3-state scenario shown in Figure~\ref{fig:MDP_picture}, between the low-CPU-usage state $s0$ and high-CPU-usage state $s2$ (becoming $s3$ after adding one state) while keeping the same structure of actions $a$'s, transition probabilities and rewards. 
We conduct the experiments on the 4-state scenario with the same transition probabilities and reward parameters as the 3-state scenario. 
The results
show the same trend as Figure \ref{fig:value_function_SDR} and Figure \ref{fig:value_function_SDN}.
The number of iterations to reach convergence is also around 60.
For the \ac{SDR} case the value functions follow $V(s0)>V(s1)>V(s2)>V(s3)$, while for the \ac{SDN} case the value functions follow $V(s3)>V(s2)>V(s1)>V(s0)$.  Besides, the optimal state-action pair selections for the 4-state case are the same as the 3-state case as well.
We are not showing the figures here due to the page limits. These results mean that our agent with value iteration is extendable to scenarios with more states and more fine-grained CPU levels.
%These results follow the same trends as the 3-state \ac{MDP} cases in the previous sections. The distances between the different state value functions (i.e. standard deviations) are similar to the 3-states results shown in Figure \ref{fig:value_function_SDR} and Figure \ref{fig:value_function_SDN} as well. 

%% file: conclusion.tex
\section{Conclusion and future work}
\label{sec:conclusion}

In this paper, we have designed and implemented an \ac{MDP}-based model to represent the dynamic CPU resource allocation problem in cloud-based wireless networks. We use value iteration algorithm to build a reinforcement learning agent to get the optimal policy. We have investigated different scenarios with different parameters as well as comparing the performance with another baseline algorithm, namely \ac{RAS}. From the simulation results we have found that our agent gets the optimal policy rapidly in under 100 iterations and the algorithm are extendable to many different scenarios. Our agent with value iteration outperforms or at least equally performs in energy savings compared to the baseline algorithm.

In this paper, we predefined the transition probabilities in the \ac{MDP} according to the statistics, which means that the agent is fully aware of the environment. In future, we plan to build more sophisticated reinforcement learning scenarios where the agent is not fully aware of the environment, and has to update its optimal decisions with time goes by. These will involve deep Q-learning techniques with neural networks.

%% file: acro_list.tex
\begin{acronym} % Give the longest label here so that the list is nicely aligned

\acro{3GPP}{Third Generation Partnership Project}
\acro{ASIC}{Application-specific Integrated Circuit}
\acro{ANN}{Artificial Neural Network}
\acro{BBU}{Baseband Unit}
\acro{C-RAN}{Cloud Radio Access Network}
\acro{CPU}{Central Processing Unit}
\acro{C-RANs}{Cloud Radio Access Networks}
\acro{D-RAN}{Distributed Radio Access Network}
\acro{DNN}{Deep Neural Network}
\acro{FDD}{Frequency Division Duplex}
\acro{FTTx}{Fibre To The x}
\acro{GPP}{General Purpose Processor}
\acro{HD}{High Definition}
\acro{IoT}{Internet of Things}
\acro{LTE}{Long Term Evolution}
\acro{LXC}{Linux Container}
\acro{MCS}{Modulation and Coding Scheme}
\acro{MDP}{Markov Decision Process}
\acro{NFV}{Network Function Vitualisation}
\acro{OFDM}{Orthogonal Frequency Division Multiplexing}
\acro{OLT}{Optical Line Terminal}
\acro{ONU}{Optical Network Unit}
\acro{OTT}{Over-the-top}
\acro{PHY}{Physical}
\acro{PON}{Passive Optical Networks}
\acro{PRB}{Physical Resource Block}
\acro{QAM}{Quadrature Amplitude Modulation}
\acro{RAS}{Random-Action-Selection}
\acro{RRH}{Remote Radio Head}
\acro{SDN}{Software-Defined Networking}
\acro{SDR}{Software-Defined Radio}
\acro{TBS}{Transport Block Size}
\acro{TDM}{Time Division Multiplexing}
\acro{vBBU}{virtual Baseband Unit}
\acro{VNO}{Virtual Network Operator}
\acro{vOLT}{vitual Optical Line Terminal}
\acro{WDM}{Wavelength Division Multiplexing}
\acro{UE}{User Equipment}

\end{acronym}